\documentclass[runningheads]{llncs}

\usepackage{microtype}
\usepackage{graphicx}
\usepackage{subcaption}
\usepackage{mwe}
\usepackage{booktabs} % for professional tables

\usepackage{hyperref}

\usepackage{amsmath}
\usepackage{amssymb}
\usepackage{mathtools}

\usepackage[]{todonotes}

\usepackage[capitalize,noabbrev]{cleveref}
\begin{document}
\title{Node Classification in Random Trees}
\author{Wouter W. L. Nuijten\orcidID{0009-0007-0689-9300} \and
Vlado Menkovski\orcidID{0000-0001-5262-0605}}
\authorrunning{W. W. L. Nuijten and V. Menkovski}
\institute{Eindhoven University of Technology, Netherlands}
\maketitle              % typeset the header of the contribution
\begin{abstract}
We propose a method for the classification of objects that are structured as random trees. Our aim is to model a distribution over the node label assignments in settings where the tree data structure is associated with node attributes (typically high dimensional embeddings). The tree topology is not predetermined and none of the label assignments are present during inference. Other methods that produce a distribution over node label assignment in trees (or more generally in graphs) either assume conditional independence of the label assignment, operate on a fixed graph topology, or require part of the node labels to be observed. Our method defines a Markov Network with the corresponding topology of the random tree and an associated Gibbs distribution. We parameterize the Gibbs distribution with a Graph Neural Network that operates on the random tree and the node embeddings. 
This allows us to estimate the likelihood of node assignments for a given random tree and use MCMC to sample from the distribution of node assignments. 

We evaluate our method on the tasks of node classification in trees on the Stanford Sentiment Treebank dataset. Our method outperforms the baselines on this dataset, demonstrating its effectiveness for modeling joint distributions of node labels in random trees.

\end{abstract}

\section{Introduction}

The node classification task is concerned with assigning a label to the nodes of a graph. 
For example, let us consider a graph where the nodes represent publications and the edges coming out of a node represent the citations of that publication. On this data structure, we can define the task of assigning the topic (from a fixed set of topics) of each publication as a node classification task. 

Typically, publications that cite each other more commonly belong to the same topic. Therefore, a topic assignment or a particular node is generally not conditionally independent from the topic assignment of its neighboring nodes. 

%For example, consider a social network where each node represents a person and the edges represent relationships between people. Each person can be assigned a label indicating their political ideology, and the task is to predict the ideology of each person based on their relationships with others. In general, the node assignments are not conditionally independent of each other, as the edges in the graph denote relations between the objects represented by the nodes. 

To achieve this property a model performing this task needs to produce a joint probability distribution over all nodes in the graph. The representation of this distribution grows exponentially in the number of nodes, and modeling this distribution from data quickly faces the curse of dimensionality. 

We utilize the topology of the graph to factorize the distribution into conditionally independent groups of variables, studied under the framework of Probabilistic Graphical Models (PGM) \cite{koller2009probabilistic}. PGMs represent individual random variables as nodes in a graph and the dependency between the two corresponding random variables as an edge. As our data structure is also a graph we can form a direct correspondence between our data and the PGM model, by assigning random variables to each node and utilizing the edges to represent the conditional dependence between the random variables. 

%In this paper we focus on a specific class, namely Markov Networks and their variations, for example, the Conditional Random Field \cite{lafferty2001conditional}.

\begin{figure}
    \centering
    \includegraphics[width=\linewidth]{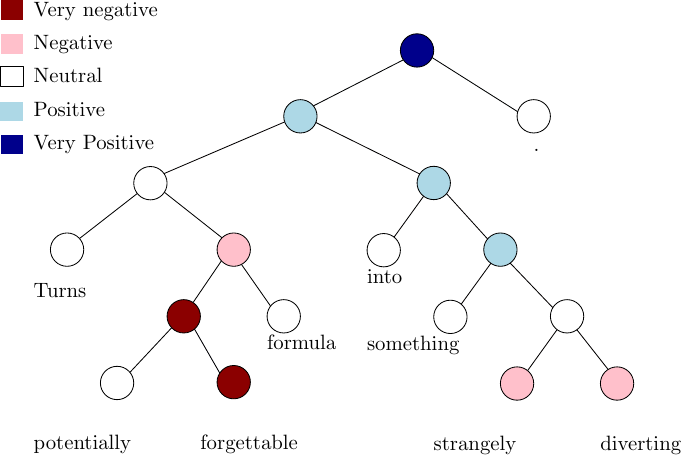}
    \caption{Example of the node classification task for sentiment analysis. The node labels and their relation encode the structural composition of sentiment.}
    \label{fig:sst_example}
\end{figure}

The task of assigning a topic to the set of publications in a broader sense is the classification of relational data, in which the entities have a number of properties and we know the structure of the relationship between the entities. Therefore, our goal is to develop models that express the relationship between the properties of each entity (or node in the graph representation), the relationships between the entities (or the edges), and the joint distribution over the class assignment to all the nodes in the graph. 

For this, a Markov Network with the corresponding graph topology is usually employed to represent the joint distribution over the class assignments. In a Markov Network, every node corresponds to a random variable, and every edge corresponds to a bidirectional dependence.

Typically, the node classification task is solved by finding the marginal distribution over class assignments using a Graph Neural Network (GNN) \cite{gori2005new,scarselli2008graph}, a model that operates on the node properties and their neighborhood in the graph. GNNs have seen considerable success in the node classification task in recent years \cite{kipf2017semisupervised,velickovic2018gat}.

When the graph structure is predetermined (i.e. all the data points share the same topology) and some of the nodes have assigned labels, determining the joint distribution over the remaining labels has been successfully implemented with a mean-field approximation parameterized by a graph neural network in \cite{bengio2020gmnn}.

In this paper, we consider the case of node classification of a set of graphs with varying topology sampled from an underlying distribution, where none of the labels are known. The mean-field approximation reduces to assuming full independence of the node assignment. This is because in a general mean-field assumption-based method we incorporate the knowledge of known labels into the distribution over unknown labels, while assuming the posterior distribution over unknown labels is factorized into independent components. If there is no knowledge about node labels available, the mean-field variational approximation has no label information to incorporate into the posterior distribution over other node labels and will therefore assume complete independence between all node labels.

To address this, our method, Neural Factor Trees, is able to combine and aggregate local information, as well as utilize the graph topology to factorize a joint probability distribution over node labels in a graph. This factorized distribution can then be sampled to obtain node labels. These samples take the dependencies between node labels into account and give more accurate solutions to the node classification task. Specifically, we produce a Gibbs Distribution that parameterizes a joint distribution over a Markov Network to find the node labels. In this paper, we implement this method to operate on trees. Markov Networks with a tree as structure, as an acyclic graph, have the benefit that inference is tractable and hence provide a way to prove the validity of our method. The method presented in this paper is hypothesized to generalize to graphs of arbitrary topology through approximate inference, however, this is out of scope for this paper. 

The main contributions of this paper are as follows:
\begin{itemize}
    \item Specification of the Node Classification task when a dataset of graphs is given and instances of the problem consist of completely unlabelled graphs (Section \ref{sec:prob_form}).
    \item Usage of a Graph Neural Network to parameterize a factor set that defines a Gibbs Distribution (Section \ref{sec:method}).
    \item Implementation of this method on a sentiment analysis dataset and empirical evidence that this method outperforms methods adopted from semi-supervised Graph Node Classification.(Section \ref{sec:evaluation}).
\end{itemize}

\section{Related work}

\subsection{Learning Probabilistic Graphical Models}
Markov Networks are trained through gradient ascent of the model parameters \cite{koller2009probabilistic} after running approximate inference using some form of belief propagation \cite{pearl1982reverend,murphy2013loopy,zhang2020factor}. Utilizing neural networks to learn the parameters of Conditional Random Fields (CRF) has also been investigated in \cite{peng2009conditional}, and although the authors regard linear chain CRFs as their main field of application, this can be generalized to arbitrary graphs. However, these methods all assume multiple draws from the same Markov Network. Since we are interested in learning a probability distribution where we witness every graph in the training set only once, the aforementioned methods can not be applied in our problem setting. However, the gradient ascent-based learning of these models inspires the idea of combining learning PGMs with other forms of gradient-based learning.

\subsection{Node classification using Graph Neural Networks}
In recent years, Graph Neural Networks have achieved significant success in semi-supervised graph node classification \cite{kipf2017semisupervised,velickovic2018gat,defferrard2016convolutional}. Graph Neural Networks are capable of learning high-dimensional representations of nodes \cite{hamilton2017inductive}. These high-dimensional node embeddings are obtained by aggregating local information between nodes to obtain and incorporate neighborhood information in node embeddings. The process of emitting and aggregating local information is called message passing \cite{gilmer2017neural}. Because they can be trained end-to-end with backpropagation \cite{rumelhart1986learning}, Graph Neural Networks can achieve state-of-the-art performance on many problems defined on graphs, such as semi-supervised node classification on the CORA \cite{sen2008collective} and SiteSeer \cite{getoor2005link} data sets. However, in their original implementation, message-passing algorithms performing graph node classification ignore dependencies between node labels because these methods assume independence between node labels. Nevertheless, the ability of Graph Neural Networks to learn high-dimensional node embeddings is a powerful property that we will leverage in order to model structural dependencies in the joint distribution. \\

There are methods that attempt to leverage the dependencies between labels in the graph \cite{wang2020unifying}. However, these methods either involve a variant of the Label Propagation algorithm \cite{zhu2005semi} or make assumptions about the correlation between label classes \cite{luan2021heterophily}. In our case, the Label Propagation algorithm is not applicable since in our problem setting there are no known labels to propagate. A promising method that does not assume prior knowledge about the correlations between the labels of neighboring vertices is the Graph Markov Neural Network method \cite{bengio2020gmnn}. Here, a Graph Neural Network is used to find a Gibbs distribution that factorizes the distribution over a Conditional Random Field \cite{bengio2020gmnn}. The authors propagate known node labels along with node features through the graph and use this to infer the unknown node labels, assuming a mean-field factorization over the unknown labels. The usage of Conditional Random Fields in \cite{bengio2020gmnn} was the inspiration to model Graph Node Classification as a Conditional Random Field in this paper, although the mean-field assumption made in this paper is less relevant in our problem setting. Nevertheless, the methods presented in \cite{bengio2020gmnn} compute a local conditional distribution of node labels, and with that, it is the method from the state-of-the-art that is most closely related to our problem formulation.\\

All things considered, there are multiple methods in the current literature that solve the (semi-supervised) node classification task. In our problem setting of having a training dataset of completely labeled graphs and having to classify previously unseen graphs, however, none of these methods is designed for this problem formulation. In the method section, we present a method that uses a Graph Neural Network to parameterize a Gibbs Distribution to solve this problem.

\section{Method} \label{sec:method}

\subsection{Problem formulation} \label{sec:prob_form}
In this paper, we study the node classification task in trees. This entails that we are given a dataset $\mathcal{D}$ of trees $\mathcal{G}_i = (V_{\mathcal{G}_i}, E_{\mathcal{G}_i})$ with node set $V_{\mathcal{G}_i}$ and edge set $E_{\mathcal{G}_i}$.
Each of the nodes $v\in V_{\mathcal{G}_i}$ in a graph has node attributes $x_v$ and a node label $y_v$. The attribute set for all nodes in a tree is denoted by $x_{V_{\mathcal{G}_i}}$, and all node labels in a graph are written as $y_{V_{\mathcal{G}_i}}$.
To reduce notational clutter, we will define the problem for a single graph and hence refer to $x_v$, $x_V$, $y_v$, and $y_V$. The problem is to infer $y_V$ given $x_V$ and the tree topology. In other words, our goal is to model the probability distribution $p(y_V \mid x_V, \mathcal{G}_i)$.

\begin{figure*}[h]
    \centering
    \begin{subfigure}[b]{0.475\textwidth}
        \centering
        \includegraphics[scale=.9]{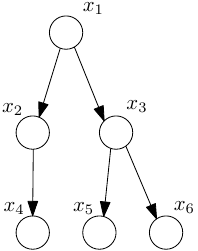}
        \caption[]%
        {{\small Input tree}}    
        \label{fig:input_tree}
    \end{subfigure}
    \hfill
    \begin{subfigure}[b]{0.475\textwidth}  
        \centering 
        \includegraphics[scale=.9]{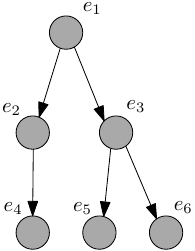}
        \caption[]%
        {{\small Node embeddings}}    
        \label{fig:node_embeddings}
    \end{subfigure}
    \vskip\baselineskip
    \begin{subfigure}[b]{0.475\textwidth}   
        \centering 
        \includegraphics[scale=.9]{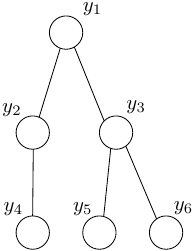}
        \caption[]%
        {{\small Markov Network}}    
        \label{fig:markov_network}
    \end{subfigure}
    \hfill
    \begin{subfigure}[b]{0.475\textwidth}   
        \centering 
        \includegraphics[scale=0.9]{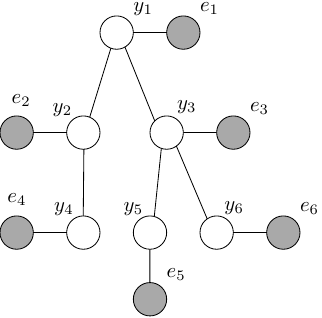}
        \caption[]%
        {{\small Conditional Random Field}}    
        \label{fig:crf}
    \end{subfigure}
    \caption[ ]
    {\small  Different graphs utilized in the construction of our method. In \ref{fig:input_tree} we see an instance of the node classification problem, and in \ref{fig:markov_network}, we see the associated Markov Network with this instance. A Graph Neural Network takes an instance of the problem and produces high dimensional node embeddings (\ref{fig:node_embeddings}). These node embeddings are combined with the Markov Network to form a Conditional Random Field (\ref{fig:crf}).}
    \label{fig:pgms_method}
\end{figure*}

In general, we can model the joint distribution over the node labels of a tree with a Gibbs Distribution that factorizes over a Markov Network.
A Gibbs Distribution is defined as a distribution $P_\Phi$ over a multivariate random variable $\mathbf{y} = \{y_1, y_2, \cdots, y_n \}$ parameterized by a set of factors $\Phi = \{\phi_1(\mathbf{D}_1), \phi_2(\mathbf{D}_2), \cdots, \phi_m(\mathbf{D}_m)\}$ if it is defined as follows:
$$ P_\Phi(y_1, y_2, \cdots, y_n) = \frac{1}{Z} \tilde{P}_\Phi(y_1, y_2, \cdots, y_n)$$
We call $\tilde{P}_\Phi(y_1, y_2, \cdots, y_n)$ the unnormalized Gibbs Measure, defined as:
$$ \tilde{P}_\Phi(y_1, y_2, \cdots, y_n) = \phi_1(\mathbf{D}_1) \times \phi_2(\mathbf{D}_2) \times \cdots \times \phi_m(\mathbf{D}_m)$$
and
$$ Z = \sum_{y_1, y_2, \cdots, y_n} \tilde{P}_\Phi(y_1, y_2, \cdots, y_n)$$
the normalization constant or partition function. The sets $\mathbf{D}$ in the case of a Markov Network are cliques in the graph, where $\bigcup_i \mathbf{D}_i = V$. In trees, this means that the sets $\mathbf{D}$ are the endpoints of the edges in the trees.
\subsection{Approach}
The challenge in learning how to parameterize the Gibbs Distribution is that the topology of the input tree in our setting is not fixed, so we cannot directly learn the factor functions. In our problem setting, the data-generating model generates node attributes along with a graph topology. This is a significant challenge in learning a Gibbs distribution in this problem setting, as the distribution varies in functional form between data points. To deal with this we develop the following method. This method is based on the idea that the Gibbs distribution over node labels is fully specified by the factor functions, so if we can estimate the factor set that best describes the underlying data generating probability distribution, we are able to solve instances of the node classification problem and obtain a more expressive distribution than if we conduct independent classification for all nodes. Therefore, we want to learn a function that takes the topology of the graph and node attributes and outputs a Gibbs Distribution that defines a joint probability distribution over node labels. Because this function has to learn the generative model that generates node embeddings and the graph topology, as well as perform Bayesian inference in this generative model, we use a Graph Neural Network (GNN) as a universal function approximator that can learn this function. To illustrate our method we also provide a visual representation of an example tree with $6$ nodes in Figure \ref{fig:pgms_method}. 

\begin{figure}[h]
    \centering
    \includegraphics[scale=0.75]{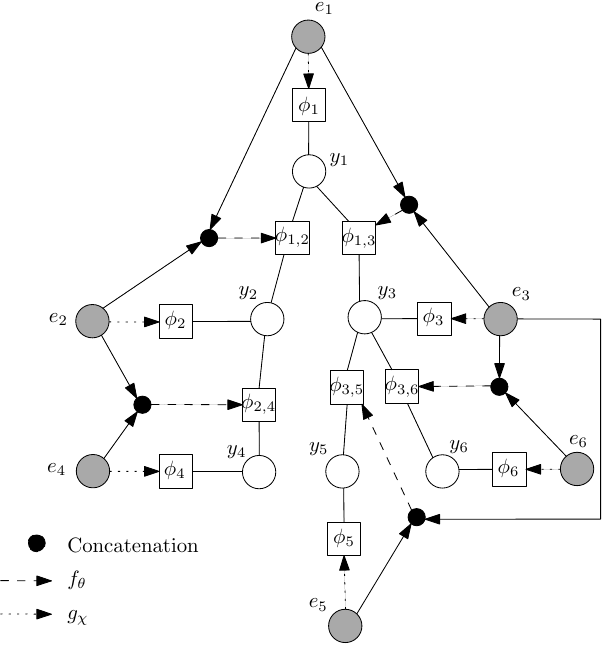}
    \caption{Factor graph construction. The node embeddings $e$ determine the node and edge factors $\phi$ through linear Neural Networks. This fully specifies a Gibbs Distribution over the resulting Markov Network. With $d$ being the number of classes a node can have and $|e|$ the size of the node embedding, $f_\theta : \mathbb{R}^{|e| + |e|} \rightarrow \mathbb{R}^{d \times d}$ is a function with trainable parameters $\theta$ that determines the edge factors from the embeddings. $g_\chi : \mathbb{R}^{|e|} \rightarrow \mathbb{R}^{d}$ is a function with trainable parameters $\chi$ that determines the node factors from the embeddings.}
    \label{fig:factor_graph}
\end{figure}

In our method, which is called Neural Factor Trees, we construct a Markov Network with the same topology as the input tree (Figure \ref{fig:input_tree} and \ref{fig:markov_network}), where the individual nodes correspond to the random variables of the Markov Network. We then use a GNN model to develop a node representation based on the node properties of the tree and its neighborhood (Figure \ref{fig:input_tree} and \ref{fig:node_embeddings}) These high dimensional node embeddings are referred to as $e_V$. We construct a CRF where each random variable is associated with an observed variable that carries the embedding produced by the GNN (Figure\ref{fig:crf}). The GNN is a key component, as it operates on graphs with various topologies and it can learn to produce a node representation that takes into account the node properties and the node's neighborhood, including the properties of the neighbors and the topology of the neighborhood. 
To specify the joint distribution over the random variables we represent the CRF as a factor graph (Figure \ref{fig:factor_graph}).

We use a linear Neural Network layer to compute the node factors from the node embeddings. The edge factors are produced by concatenating the embeddings of the endpoint nodes and passing this vector through a linear Neural Network layer. The embedding of the parent node here comes first, in order to keep a consistent permutation scheme.

\subsection{GNN design} 
For the GNN model, we use a Message Passing Neural Network (MPNN) and specifically the Gated Graph Neural Networks model \cite{li2015gated}. In this algorithm, we use a Multilayer Perceptron to calculate the messages emitted by all vertices. Note that the official algorithm in \cite{li2015gated} restricts this Multilayer Perceptron to a single layer, whereas we do not impose this constraint. We found that using a transfer function with 2 layers in the Multilayer Perceptron provided us with a slight improvement in our model evidence. This design choice was originally made to make our GNN unit more expressive. In every node, we collect and sum incoming messages and update node representations using a Gated Recurrent Unit \cite{cho2014properties}. The update rules are as follows:
 \begin{equation}
     \begin{aligned}
      h_V^0 = \left[\mathbf{X}_V \mid \mid  \mathbf{0} \right] \\
      a_i^t = \sum_{j \in \mathcal{N}(i)} f_\theta( h_j^t) \\
      h_i^{t+1} = \textrm{GRU}(a_i^t, h_i^t)
     \end{aligned}
 \end{equation}
 Here, $||$ is the concatenation operator. Because we use a Gated Recurrent Unit (GRU) \cite{cho2014properties} as an update function, every vertex is able to distill relevant information from incoming messages and update its representation accordingly.
\par
As this method only utilized Gated Graph Neural Network and linear layers, the model is fully differentiable with respect to its parameters. As is common in training Probabilistic Graphical Models, we optimize the log-likelihood of the produced distribution given a set of samples from the true distribution $\mathcal{D}$. The log-likelihood is a function of the produced factor set $\Phi$, which in our case is determined by the model $p_\psi\left(\Phi \mid \mathbf{G}_{\mathcal{D}_i}, \mathbf{X}_{\mathcal{D}_i}\right)$. Here, to avoid visual clutter, we let $\Phi_\psi$ be shorthand for $p_\psi\left(\Phi \mid \mathbf{G}_{\mathcal{D}_i}, \mathbf{X}_{\mathcal{D}_i}\right)$, and $Z_{\Phi_\psi}$ for the normalization constant of the Gibbs Distribution produced by $\Phi_\psi$.
\begin{equation} \label{eq:log-likelihood-functional}
    \mathcal{L}(\psi : \mathcal{D}) = \sum_{i=1}^{|\mathcal{D}|}\sum_{\phi \in \Phi_\psi} \left( \log \phi(\mathbf{D}_{\phi})\right) - \log Z_{\Phi_\psi}
\end{equation}
Here, $\psi$ is the set of all trainable parameters of the model, including $\theta$ and $\chi$ in figure \ref{fig:factor_graph}.
As we have trees in our problem statement, the normalization constant is tractable to compute \cite{koller2009probabilistic}.
With this setup, we have defined a fully differentiable method and an adequate quality metric with which to train our method. 
\subsection{Classifying nodes} 
With the produced Gibbs Distribution we can construct solutions to the node classification problem. In order to do this we utilize a Markov Chain Monte Carlo (MCMC) \cite{hastings1970monte} algorithm to produce samples from the produced probability distribution. We utilize the Gibbs Sampling \cite{geman1984stochastic} algorithm. The advantage of the Gibbs Sampling algorithm is that it produces unbiased samples from the distribution on which it operates. The disadvantage of Gibbs Sampling, as with most MCMC algorithms, is that it is computationally expensive to run. Alternatively, because we only consider trees in this paper, we can use the Max-Product algorithm to obtain the mode of the Gibbs Distribution. However, since this research is inspired by a desire to model the full distribution over node labels and is therefore also interested in other probable assignments of node labels, we use the Gibbs Sampling algorithm to obtain a histogram of probable node label assignments.

\section{Evaluation} \label{sec:evaluation}
\subsection{Dataset}
The Stanford Sentiment Treebank (SST) \cite{socher2013recursive} dataset is a dataset of fully labeled parse trees of sentences from movie reviews. Each parse tree consists of a single sentence and every node of the parse tree is labeled by its sentiment (Figure~\ref{fig:sst_example}). The combination of nodes in the tree allows for the analysis of compositional relations of sentiment in language. The dataset consists of 11855 fully labeled sentences, where every node gets a sentiment score out of 5 categories: very negative, negative, neutral, positive, and very positive. The task is to classify all nodes in each tree. Due to the compositional relations of sentiment in these sentences, we cannot assume conditional independence of the label assignment.

We use GloVe \cite{pennington2014glove} word embeddings as node attributes $x_v$. The unlabeled version of this graph corresponds to Figure \ref{fig:input_tree} of our method. We use GloVe embeddings to isolate the performance of our method and not rely on the word embeddings we learn as a by-product of our method. The GloVe embeddings are not to be confused with the node embeddings our method produces after a pass through its GNN.

\subsection{Experiments}
The performance of our method can be evaluated in multiple ways; a theoretical and an empirical performance metric. Since we can compute the log-likelihood of a label configuration for the joint probability distribution, we have access to a theoretical performance metric. Finding the maximum likelihood estimate of a Gibbs Distribution in trees can be done with the Max-Product algorithm. Furthermore, we can sample the produced Gibbs Distribution and provide the empirical mode of the sample distribution. This provides us with a maximum likelihood estimate of the label assignments of the tree under our produced Gibbs Distribution and we are therefore able to compute the accuracy of our method. \\

We test the performance of our method\footnote{The code for these experiments is available on \href{https://github.com/wouterwln/NeuralFactorTrees}{Github} (\url{https://github.com/wouterwln/NeuralFactorTrees}).}  against a base-line  Graph Neural Network model that assumes full independence of the class label of the nodes, as well as a Graph Markov Neural Network (GMNN) \cite{bengio2020gmnn}, which uses a mean-field approximation of the distribution over node labels. In the related work GMNN has been identified as a state-of-the-art method for semi-supervised graph node classification that is closely related to our problem formulation.
However, since there is no information about node labels available during inference, we expect this mean-field approximation to reduce to independent classification. Nevertheless, because of its effectiveness in computing local conditional distributions, we use GMNN as a baseline method to test our method against. A disadvantage of our method compared to GMNN is that the maximum a posteriori assignment of labels under the posterior distribution in GMNN is easily found, as it is the node-wise maximum a posteriori estimate of the categorical distributions in every node. For our method, we have to employ the Max-Product algorithm to find the maximum a posteriori estimate or report the mode of the empirical dataset sampled from the Gibbs Distribution, which is significantly more computationally expensive. 

\begin{table}
    \centering

    \begin{tabular}{c|c c c}
     & GNN & GMNN & Neural Factor Trees  \\ \hline
    Log-likelihood & $-0.522$& $-0.515$& $\mathbf{-0.469}$ \\
    Accuracy & $73.9\%$ & $74.0\%$ & $\mathbf{78.3\%}$
\end{tabular}
    \caption{Results on the SST Dataset}
    \label{tab:sst_dataset_results}
\end{table}
\vspace{-1cm}
\subsection{Results}
In Table \ref{tab:sst_dataset_results} we display the results of our method on the SST dataset. The log-likelihoods presented in this table are aggregated over the number of nodes in a tree, as a large tree would otherwise have a smaller log-likelihood due to the individual contributions by the nodes. We see that the data has a higher log-likelihood under the distributions we produce than results from both independent classification using Graph Neural Networks and GMNN methods, which means that the learned distributions better represent the dataset. We also see this reflected in the accuracy scores. An interesting detail is that GMNN does not outperform independent classification by a significant margin, supporting our hypothesis that the mean-field assumption reduces to independent classification when none of the labels are known.\\

\section{Conclusion}

In this paper, we described a novel way of addressing local dependencies in graph labels when conducting random tree node classification. We have developed the Neural Factor Trees method; a method that is able to take the topological properties of the graph into consideration and can produce a joint probability distribution over node labels. Our method can be trained end-to-end using backpropagation. We show that our method is able to outperform existing methods on the SST Dataset. \\

The method described in this paper operates on trees, and this property is leveraged in the computation of the log-likelihood. The fact that the graphs in the dataset are trees ensures the computation of the normalization constant is tractable in linear time. If this tree property is dropped, we are no longer guaranteed to be able to compute the normalization constant in polynomial time. This is a known problem in learning Probabilistic Graphical Models and might also be a hurdle when generalizing our method to non-tree graphs. We, therefore, identify this as the main limitation of our method, as this tree property is not always satisfied. Furthermore, we only model the dependencies between node labels of nodes that are connected in the input tree. However, these (conditional) dependencies might exist for more than 1 step in the graph. With our current method, we are not able to model these interactions, instead depending on the dependency between labels of two distant nodes flowing over the path between these nodes in the graph.  \\

An extension of this work might investigate an alleviation of these drawbacks and might investigate the performance of our method under approximate inference (e.g. belief propagation) for the computation of the partition function. This would extend our method to operate on any graph instead of trees, and the interesting research avenue is how our method performs under different forms of approximate inference.

\bibliographystyle{splncs04}
\bibliography{bibliography}
\end{document}